%% file: main.tex
\documentclass[conference]{IEEEtran}
\IEEEoverridecommandlockouts
\usepackage{cite}
\usepackage{amsmath,amssymb,amsfonts}
\usepackage{graphicx}
\usepackage{textcomp}
\usepackage{xcolor}
\usepackage{url}
\usepackage{graphicx}
\usepackage{tabularx}
\usepackage{subcaption}
\usepackage{algpseudocode}
\usepackage{xspace}
\usepackage{tabularx}
\usepackage{bm}
\usepackage{multirow}
\usepackage{enumitem}
\usepackage[ruled,vlined,linesnumbered]{algorithm2e}
\usepackage{color,soul}
\usepackage{wrapfig}
\usepackage{amsmath,multicol, mathrsfs}
\usepackage[bottom]{footmisc}
\usepackage{makecell}

\newcommand{\ourdata}{{SMILE-College}}

\makeatletter

\def\BibTeX{{\rm B\kern-.05em{\sc i\kern-.025em b}\kern-.08em
    T\kern-.1667em\lower.7ex\hbox{E}\kern-.125emX}}
\begin{document}

\title{Understanding Student Sentiment on Mental Health Support in Colleges Using Large Language Models}

\author{\IEEEauthorblockN{Palak Sood, Chengyang He, Divyanshu Gupta, Yue Ning, Ping Wang}
\IEEEauthorblockA{\textit{Department of Computer Science} \\
\textit{Stevens Institute of Technology}\\
Hoboken NJ, USA \\
{\{psood, che14, dgupta12, yue.ning, pwang44\}@stevens.edu}}

}

\maketitle

\begin{abstract}

Mental health support in colleges is vital in educating students by offering counseling services and organizing supportive events. However, evaluating its effectiveness faces challenges like data collection difficulties and lack of standardized metrics, limiting research scope. Student feedback is crucial for evaluation but often relies on qualitative analysis without systematic investigation using advanced machine learning methods. This paper uses public Student Voice Survey data to analyze student sentiments on mental health support with large language models (LLMs). We created a sentiment analysis dataset, SMILE-College, with human-machine collaboration. The investigation of both traditional machine learning methods and state-of-the-art LLMs showed the best performance of GPT-3.5 and BERT on this new dataset. The analysis highlights challenges in accurately predicting response sentiments and offers practical insights on how LLMs can enhance mental health-related research and improve college mental health services. This data-driven approach will facilitate efficient and informed mental health support evaluation, management, and decision-making.
\end{abstract}


\input{1-Introduction}

\input{2-Dataset}
\input{3-Methods}

\input{4-Experimentation}

\input{5-Conclusion}

\section*{Acknowledgment}
This work was supported in part by the US National Science Foundation grants IIS-2245907,  2047843, and 2437621, and the startup funding from the Stevens Institute of Technology. 

\bibliographystyle{IEEEtran}
\bibliography{sample-base}

\end{document}

%% file: 1-Introduction.tex
\section{Introduction}

Mental health has become a paramount concern within the student community, increasingly recognized as essential to both their overall well-being and academic success~\cite{mentalhealth13,LIPSON2022138}.
A 2020 report by the National Institute of Mental Health highlights that mental illness prevalence is highest among those under 25 years, including 67\% of college students~\cite{pester2023importance}. Universities play a crucial role by offering counseling services and organizing events to support students' emotional well-being. However, evaluating mental health support in colleges faces challenges such as data collection difficulties, lack of standardized metrics, insufficient funding, and limited inter-institutional collaboration~\cite{mentaldata09,investmental19}. These issues restrict research scope, with most studies \cite{doi:10.1080/03069885.2012.743110,articleHMS, articleHMS2} focusing on student mental health status rather than the effectiveness of support services.

Student feedback is vital for assessing university mental health services. Surveys like the Healthy Minds Study \cite{healthymindsstudy} and the American College Health Association Health Assessment \cite{ACHA} gather insights into students' mental health and service utilization. Universities can use this feedback to improve their initiatives. Recent studies have explored student perspectives on mental health support~\cite{priestley2022student,osborn2022university}, but key challenges remain. These include reliance on qualitative analysis from limited feedback, lack of comprehensive quantitative evaluations, and the absence of utilizing advanced machine learning methods to analyze sentiments. Additionally, existing datasets do not support developing machine learning models for this purpose. 

\begin{table*}[t]
\centering
  \caption{Representative examples for each sentiment category.}
  \begin{tabular}{l|p{0.75\linewidth}}
    \hline
    \bf Label & \bf Students' Survey Response \\
    \hline
    \bf Satisfied & I honestly think all of it is \textcolor{red}{amazing so far}, I visit the therapists and nurses a lot right now and it’s \textcolor{red}{all been covered by tuition and fees. everyone is super friendly} and I always leave feeling like I had \textcolor{red}{everything taken care of} \\\hline
    \bf Dissatisfied & I \textcolor{blue}{only know of one mental health employee} but \textcolor{blue}{not know how to reach them or what to do} . the therapy they provide is \textcolor{blue}{also catholic based which I am not} \\\hline
    \bf Mixed & my college \textcolor{red}{works well} in communicating about the various mental health resources on campus. \textcolor{blue}{more attention is needed} to expand the mental health department in its diversity.  \\\hline
    \bf Neutral & \textcolor{green}{I haven't personally used any} of the services, so I feel as though \textcolor{green}{I am not qualified} to answer this question.  \\\hline
\end{tabular}
\label{tab:sentiment_examples}
\vspace{-4mm} 
\end{table*}

This paper aims to utilize the Student Voice Survey (SVS) data by College Pulse \cite{collegepulse} to create a sentiment analysis dataset and explore the potential of large language models (LLMs) for predicting students' sentiments.
Specifically, we utilize the students' narrative feedback in SVS data about their feedback for the advantages and disadvantages of mental health support in their colleges. 
To create the dataset for sentiment analysis, 
we first explore the spectrum of students' sentiments by leveraging the power of LLMs, considering the standard three categories of sentiment labels (including ``Satisfied'', ``Dissatisfied'', and ``Neutral''), and designing a task-specific coarse prompt. 
This investigation motivates us to adopt a more detailed analysis by introducing a new sentiment category ``Mixed''.
With the more nuanced set of sentiment categories, we collect the sentiment labels of students' responses in SVS data with human annotation, validation, and collaborative discussion.

The enriched SVS data, named \textbf{S}enti\textbf{M}ent analys\textbf{I}s of students' menta\textbf{L} h\textbf{E}alth support in \textbf{College}s (\ourdata), includes 793 records with sentiment labels and is publicly available online\footnote{\url{https://github.com/LEAF-Lab-Stevens/SMILE-College}}.
Representative examples for each category are shown in Table~\ref{tab:sentiment_examples}.
We aim to investigate three tasks:
(1) \textit{Sentiment prediction}: Automatically predicting sentiment labels by designing task-specific prompts for LLMs with fine-grained sentiment categories.
(2) \textit{Prediction error analysis}: Analyzing LLM prediction errors across sentiment categories.
(3) \textit{Support limitation identification}: Using LLMs, embedding learning, and clustering techniques to identify key limitations of mental health support based on ``Dissatisfied'' responses.
To the best of our knowledge, this is the first work to comprehensively evaluate student mental health support in colleges from students' perspective. This data-driven study enables the automatic prediction of students' perceptions of mental health support with advanced LLMs, providing quantitative and qualitative assessment.

In summary, our study makes the following \textbf{key contributions}. 
\begin{itemize}
    \item Created the first sentiment analysis dataset of student mental health support in colleges by annotating sentiment labels with human-machine collaboration. 
    \item Investigated several state-of-the-art LLMs on the \ourdata\ data for three important sentiment prediction-related tasks with a fine-grained prompt.  
    \item Experimental results highlight the better performance of GPT-3.5 and BERT on this specific task and underscore the challenges in accurately predicting the response sentiments.
    \item Identified key limitations of mental health support for potential improvement in colleges by leveraging the power of LLMs, embedding learning, and clustering techniques. 

\end{itemize}

\section{Related Work}

The significance of mental health within student communities has escalated, underscoring the essential role of support services. 
Existing research about mental health in colleges mostly focuses on investigating students' mental health status \cite{doi:10.1080/03069885.2012.743110,articleHMS, articleHMS2}.
Research on evaluating mental health support in colleges is limited due to various challenges. 
Various surveys, such as the Student Voice Survey and the Healthy Minds Study \cite{healthymindsstudy}, are designed to gain students' insights on mental health services for assessing these services.
The American College Health Association also conducted a survey to collect students' perceptions, behaviors, and habits \cite{ACHA,articleACHA, articleACHA2, articleACHA3}. 
There are some recent works about student perspectives on improving mental health support services in universities or systematically reviewing the students' use of mental health services in universities \cite{priestley2022student,osborn2022university}. 
However, there are still several key challenges that have not been addressed, such as qualitative analysis on limited feedback, lack of comprehensive quantitative evaluation, and the absence of utilizing advanced machine learning methods for the evaluation.

Sentiment analysis is the computational study of opinions, attitudes, and emotions expressed in narrative texts \cite{liu2022sentiment}.  
Deep learning models, including recurrent neural networks and transformer-based models, have been successfully employed in sentiment analysis \cite{Ravi2015ASO, zhang2016text,wang-etal-2016-attention,sood2023enhancing}, while lexicon- and rule-based methods relying on sentiment dictionaries have also been widely used \cite{Taboada2011LexiconBasedMF}.  
Sentiment analysis has also been applied to social media data for the detection of signs of depression and suicidal ideation, as demonstrated by Shen et al.~\cite{Shen2017DepressionDV}. 
Recently, the pre-trained and large language models gained significant attention in the field of sentiment analysis~\cite{howard-ruder-2018-universal}.
Sentiment analysis has been widely applied to various applications, such as social media \cite{9702359} and customer feedback~\cite{9395802}, and demonstrated its effectiveness.  
It has also been used in various mental health prediction and analysis tasks \cite{guo2024large}. 
Most studies focus on analyzing individuals' mental health by examining their emotional sentiments, leveraging sentiment analysis to understand mental health states or detect early signs of mental health disorders \cite{benrouba2023emotional,shah2020mental}.

However, to the best of our knowledge, no prior work has explored evaluating students' perceptions of mental health support in colleges using sentiment analysis. The lack of such research leaves a critical gap in understanding how students perceive and engage with the mental health support resources available to them in colleges, which is essential for developing a more effective, student-centered mental health support system for each college.  
This paper fills this gap by introducing the first sentiment analysis dataset with a specific focus on student perceptions of mental health support in college settings. 
By creating and analyzing the dataset with LLMs, this work provides a foundation for data-driven evaluation and decision-making and offers insights into students' satisfaction and concerns based on their experiences with available mental health services.
More importantly, the dataset will not only support related research into the sentiment analysis associated with mental health support but also has the potential to identify support limitations for actionable strategies to tailor mental health support to students' real needs.

%% file: 2-Dataset.tex
\section{The SMILE-College Dataset}
\label{sec:dataset}
To the best of our knowledge, no existing dataset has been specifically developed for sentiment analysis on mental health support in colleges.
This section provides the details of data creation of the 
SMILE-College dataset for sentiment analysis.

\begin{figure*}[!tp]
    \centering
    \includegraphics[width=1.0\textwidth]{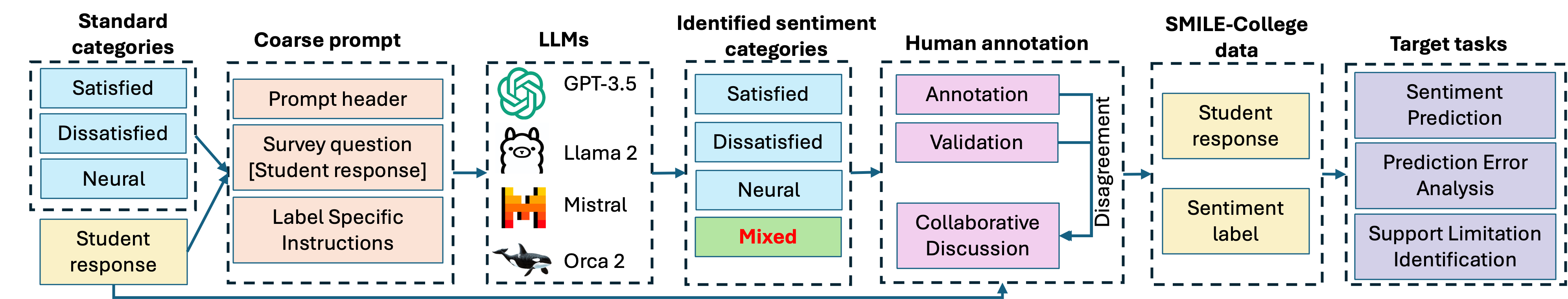}
    \caption{Overall framework for sentiment annotation with human-machine collaboration.}
    \label{fig:overall_framework}
\end{figure*}

\subsection{\textbf{Student Voice Survey (SVS) Data}} 
\label{sec:data_svs}

This study uses the SVS response data\footnote{\url{https://reports.collegepulse.com/current-state-of-mental-health}} on the current state of mental health designed by College Pulse, to examine the social and emotional well-being of students and gain insights into their attitudes, preferences, and behaviors.
The survey, conducted in 2022, comprised 20 questions and was completed by 2,000 undergraduate students from a panel representing over 1,500 colleges and universities across the United States. 
Our study focuses on the text responses to the question ``\textbf{\textit{What mental health or wellness services and supports provided by your college are working well? What aspects of mental health and wellness need more attention?}}''


Out of the 2,000 response records, 202 responses represented null values like ``n/a", ``na", ``none", ``idk" and empty value. There were also many one word entries that did not provide meaningful answers as shown in first category in Table \ref{tab:examples_of_bad_data}.  Additionally, several comments only mentioned the name of a service without providing detailed feedback or indicating satisfaction levels (second category of Table \ref{tab:examples_of_bad_data}). Also, many responses were overly brief and ambiguous in conveying satisfaction or dissatisfaction as seen in the Irrelevant category, thus impacting the quality of data annotation and model prediction. 
To ensure the quality of the dataset, we removed this data, by setting a minimum word count of 12 words (based on manual assessment) and removing the irrelevant records during annotation. After this refinement process, we obtained a condensed dataset of 793 records with sufficient context for sentiment analysis.

\subsection{\textbf{Sentiment Annotation with Human-Machine Collaboration}}
\label{sec:data_annotation_SMILE-College}
Based on the selected data samples, we annotate sentiment labels in a human-machine collaborative manner. While human annotation ensures high accuracy and nuanced understanding, it is costly and time-consuming. Moreover, manually analyzing many samples and defining the appropriate range of sentiment categories becomes challenging. 
Recently, LLMs offered a scalable solution for annotation \cite{tan2024large}. However, there is a significant reduction in performance when transitioning from human labels to LLMs' generated labels due to the inherent noise in the generated labels \cite{pmlr_v239_mohta23a,ziems2024can}.
Therefore, a viable alternative is to have humans and LLMs work together on this specific annotation task \cite{wang_etal_2021_want_reduce}.


During our annotation, both LLMs and human annotators contributed unique strengths in a complementary twofold approach.
LLMs were first used for quick preliminary analysis, facilitating the identification of sentiment patterns across the entire dataset. By leveraging multiple LLMs, we detected edge cases that suggested the need for an additional sentiment category, enhancing the dataset's granularity. 
LLMs also helped filter responses with irrelevant or insufficient information, streamlining the annotation process and improving the overall efficiency.
Meanwhile, human annotators brought essential depth and contextual understanding of the sentiments, particularly in cases where nuanced interpretation was required. 
Together, this human-machine collaboration strategy ensured the accuracy and consistency in sentiment annotation and enabled a robust, context-sensitive dataset for analyzing sentiment on mental health support in colleges.
The sentiment annotation of \ourdata\ can be summarized as the following three steps.
Figure \ref{fig:overall_framework} provides the overall framework of the annotation.


\begin{table}[!tp]
\centering
\caption{Examples of survey records that were removed from the dataset.}
\begin{tabularx}{\linewidth}{l|X}
    \hline
    \bf Category & \bf Examples \\
    \hline
    \textbf{Non-seriousness} & 
    \begin{itemize}[leftmargin=*]
        \setlength{\itemsep}{0pt}
        \vspace{-1mm}
        \item Nsvdejsj 
        \item Unknown 
        \vspace{-1mm}
    \end{itemize} \\
    \hline
    \textbf{Insufficient Information} & 
    \begin{itemize}[leftmargin=*]
        \setlength{\itemsep}{0pt}
        \vspace{-1mm}
        \item eating disorders!!!
        \item tutoring, counseling, and professor care
        \vspace{-1mm}
    \end{itemize} \\
    \hline
    \textbf{Irrelevant Information} & 
    \begin{itemize}[leftmargin=*]
        \setlength{\itemsep}{0pt}
        \vspace{-1mm}
        \item while I was home, I felt that school was not worth it as I was home and not doing or going anywhere.
        \item yeah, I'll be there at around noon, and I just got home and I’ll get back home from church lol I have a lot of stuff going to my house so I’ll
        \vspace{-1mm}
    \end{itemize} \\
    \hline
\end{tabularx}
\label{tab:examples_of_bad_data}
\end{table}

\begin{table*}[!tp]
\centering
\caption{Statistics of the SMILE-College dataset.}
\begin{tabular}{l|cccc}
\hline
&\bf Satisfied &\bf Dissatisfied &\bf Mixed &\bf Neutral \\ \hline
No. of Records & 107 & 376 & 220 & 90 \\

Average response length (in words) & 21.84 & 33.01 & 28.06 & 18.15 \\
Min response length (in words) & 12 & 12 & 12 & 12 \\
Max response length (in words) & 93 & 199 & 106 & 46 \\
Average \# of sentences in responses & 1.89 & 2.51 & 2.42 & 2.03 \\
Min \# of sentences in responses & 1 & 1 & 1 & 1 \\
Max \# of sentences in responses  & 7 & 11 & 8 & 5 \\

\hline
\end{tabular}
\label{tab:data_stats}
\end{table*}

\textit{Step 1. Sentiment Annotation with LLMs.} 
Initially, the number of categories in our data was unclear. To navigate the unstructured nature of the survey responses, we employed Large Language Models (LLMs) to identify response clusters. The goal was to classify the responses into three standard categories: Satisfied (positive class), Dissatisfied (negative class), and Neutral (neutral class).
To achieve this, we designed and refined a prompt-engineered approach, leveraging the advanced linguistic capabilities of LLMs. Our strategy involved creating a \textbf{coarse prompt} that consisted of three key components:
\begin{itemize}
    \item Prompt Header: This section contained task-specific instructions, guiding the LLMs to adopt the role of an experienced analyst specializing in mental health text analysis. Here is how we assigned the role in the prompt: ``\textit{You are a very experienced analyst trying to analyze the answers to a question asked during a mental health survey. No answer will explicitly mention any of the categories. You have to analyze them based on the rules and categorize them in one word, SATISFIED, DISSATISFIED, or NEUTRAL.}''
    
    \item Survey Question and Student Response: This component ensured that the LLMs' evaluation was directly informed by the specific content of the survey, grounding its sentiment analysis in the precise context of the student's responses. The survey question is provided in Section \ref{sec:data_svs}.
    
    \item Label-Specific Instructions: Comprehensive guidelines were provided for each sentiment category, facilitating accurate categorization of the sentiment expressed in the responses. Guidelines were similar to the criteria for human annotation and validation of sentiment labels in Step 3 of this section.
\end{itemize}

This zero-shot learning strategy, supported by the coarse prompt, allowed for a preliminary exploration of the spectrum of students' sentiments towards mental health services. By adopting this approach, we were able to harness the capabilities of LLMs to effectively categorize sentiments, despite the initial ambiguity regarding the number of categories.

\textit{Step 2. Sentiment Category Identification. }
We investigated multiple LLMs using the coarse prompt and found a 50.15\% prediction agreement for all the models. Among the agreed records, none were classified as neutral, and 70.7\% were classified as dissatisfied. 
Based on our manual review of the predictions and the cases where model outputs diverged, {we observed that a significant proportion of the responses contained both positive and negative sentiments, making it difficult to classify them into one of the standard sentiment categories. For example, consider the response ``\textit{The student support team at my college has been very helpful in supporting my mental health. I think more attention should be paid on the negative impact of stress from school has on students' mental health.}" Here, the first sentence expresses satisfaction, while the second reflects dissatisfaction.} This blend of sentiments highlights the need for a more nuanced annotation approach, as traditional three sentiment categories may not fully capture the complexity of responses in this context.

Inspired by this insight, we shift to a more detailed analysis by introducing ``\textit{Mixed}" as a new sentiment category. This transition marks a significant change from a broad to more nuanced sentiment analysis, allowing for a deeper and more precise understanding of the survey responses through human-machine collaboration.

\textit{Step 3. Human Annotation and Validation of Sentiment Labels. } 
Based on the four categories of students' sentiments identified in Step 2, we adopt a {two-stage human annotation process}. The preliminary annotation stage is executed by one annotator, which is subsequently subjected to a validation stage conducted by another annotator. Both annotators are graduate students who are proficient in English. 
This rigorous process revealed a disagreement rate of 9.98\% between the annotations in two phases. These mismatches are resolved collaboratively through collective discussions among the annotators and other researchers involved, leading to a consensus on the final sentiment labels.

\label{sec:rules}
The specific criteria for the annotation of each category are as follows. 
(1) ``\textit{Satisfied}": at least 75\% of the language expressed satisfaction, with minimal suggestions for improvement. 
(2) ``\textit{Dissatisfied}": at least 75\% of the language indicated discontent or suggestions for enhancement, with little mention of satisfaction. 
(3) ``\textit{Mixed}": expressions of satisfaction and dissatisfaction/suggestions were approximately evenly split, with each constituting about 50\%. 
(4) ``\textit{Neutral}": no clear emphasis on satisfaction, dissatisfaction, or suggestions for improvement. This discourse focuses on mental health in a college context.

This human-machine collaboration annotation strategy not only enhances this specific sentiment analysis task but also highlights the importance of combining computational analysis with human insights to capture the intricate emotional nuances within students' responses. 
We enrich the SVS data with the sentiment labels and obtain the SMILE-College dataset for sentiment analysis.

\textbf{SMILE-College Data Statistics.}
Table \ref{tab:sentiment_examples} provides one representative example of each sentiment category. Table \ref{tab:data_stats} illustrates the basic data statistics of the SMILE-College data. Following data filtering procedures, 266 distinct colleges/universities were covered within the dataset. Notably, the word count distribution across records ranges from a minimum of 12 words to a maximum of 199 words whereas the sentence count ranges from 1 to 11 sentences. 

\subsection{\textbf{Target Tasks}}

To evaluate the usability of the \ourdata\ data, we investigated three important tasks, including: 
\begin{itemize}
    \item \textit{Sentiment prediction} (\textbf{T1}): text-based multi-class classification of sentiment labels for students' responses with a task-specific fine-grained prompt for LLMs. 
    \item \textit{Prediction error analysis} (\textbf{T2}): examine the prediction errors of LLMs across different sentiment categories. 
    \item \textit{Support limitation identification} (\textbf{T3}): based on the responses labeled as ``Dissatisfied'', we utilize the capabilities of LLMs, embedding learning, and clustering techniques to pinpoint the main shortcomings in student mental health support in colleges. 
\end{itemize}



%% file: 3-Methods.tex
\begin{table*}[!tp]
\centering
\caption{Overall sentiment prediction performance and detailed breakdown of predictions across each sentiment category on the SMILE-College \textit{test set}.}
\resizebox{\linewidth}{!}{
\begin{tabular}{l|ccc|ccc|ccc|ccc|ccc}
\hline
 & \multicolumn{3}{c|}{\textbf{Satisfied}} & \multicolumn{3}{c|}{\textbf{Dissatisfied}} & \multicolumn{3}{c|}{\textbf{Mixed}} & \multicolumn{3}{c|}{\textbf{Neutral}} & \multicolumn{3}{c}{\textbf{Overall}} \\
\hline
& \textbf{Precision} & \textbf{Recall} & \textbf{F1} & \textbf{Precision} & \textbf{Recall} & \textbf{F1} & \textbf{Precision} & \textbf{Recall} & \textbf{F1} & \textbf{Precision} & \textbf{Recall} & \textbf{F1} & \textbf{Precision} & \textbf{Recall} & \textbf{F1} \\
\hline
\textbf{LR} & 0.45 & 0.24 & 0.31 & 0.64 & 0.77 & 0.70 & \underline{0.59} &  0.55 & 0.57 & 0.69 & 0.58 & 0.63 & 0.61 & 0.62 & 0.61 \\

\textbf{SVM} & 0.50 & 0.33 & 0.40 & 0.63 & \underline{0.78} & 0.70 & 0.53 & 0.40 & 0.46 & 0.71 & 0.63 & 0.67 & 0.60 & 0.61 & 0.60 \\

\textbf{BERT} & 0.68 & \underline{0.71} & 0.70 & 0.88 & \textbf{0.85} & \textbf{0.86} & \textbf{0.62} & 0.70 & \underline{0.66} & \underline{0.88} & 0.74 & \underline{0.80} & \underline{0.79} & \underline{0.78} & \underline{0.78} \\

\textbf{Mistral} & \underline{0.83} & 0.48 & 0.61 & \underline{0.98} & 0.52 & 0.68 & 0.56 & 0.50 & 0.53 & 0.28 & \textbf{1.00} & 0.43  & 0.77 & 0.57 & 0.60 \\
\textbf{Orca 2} & \textbf{0.88} & 0.61 & \underline{0.72} & 0.95 & 0.25 & 0.40 & 0.34 & \textbf{0.95} & 0.50 & \textbf{1.00} & 0.17 & 0.29 & 0.78 & 0.48 & 0.46  \\
\textbf{Llama 2 } &  0.00 & 0.00 & 0.00 & 0.93 & 0.63 & \underline{0.75}  & 0.50 & \textbf{0.95} & \underline{0.66} & 0.52 & 0.79 & 0.62 & 0.65 & 0.65 & 0.61 \\

\textbf{GPT-3.5} & 0.66 & \textbf{0.90} & \textbf{0.76} & \textbf{1.00} & 0.75 & \textbf{0.86} & \textbf{0.62} & \underline{0.78} & \textbf{0.69} & 0.81 & \underline{0.89} & \textbf{0.85}  &  \textbf{0.84} & \textbf{0.79} & \textbf{0.80} \\

\hline
\end{tabular}
}
\label{tab:results-smile-data}
\end{table*}

\begin{table*}[!tp]
\centering
\caption{Overall sentiment prediction results on \textit{the entire SMILE-College dataset} with zero-shot prompting using LLMs.}
\resizebox{\linewidth}{!}{
\begin{tabular}{l|ccc|ccc|ccc|ccc|ccc}
\hline
 & \multicolumn{3}{c|}{\textbf{Satisfied}} & \multicolumn{3}{c|}{\textbf{Dissatisfied}} & \multicolumn{3}{c|}{\textbf{Mixed}} & \multicolumn{3}{c|}{\textbf{Neutral}} & \multicolumn{3}{c}{\textbf{Overall}} \\
\hline
& \textbf{Precision} & \textbf{Recall} & \textbf{F1} & \textbf{Precision} & \textbf{Recall} & \textbf{F1} & \textbf{Precision} & \textbf{Recall} & \textbf{F1} & \textbf{Precision} & \textbf{Recall} & \textbf{F1} & \textbf{Precision} & \textbf{Recall} & \textbf{F1} \\
\hline

\textbf{Mistral} & 0.88 & 0.28 & 0.43 & \underline{0.95} & 0.46 & 0.62 & \underline{0.61} & 0.64 & 0.62 & 0.26 & \textbf{0.99} & 0.41 & 0.77 & 0.54 & 0.57 \\
\textbf{Orca 2} & \underline{0.89} & \underline{0.56} & \underline{0.69} & \textbf{0.96} & 0.23 & 0.37 & 0.32 & \textbf{0.97} & 0.48 & \textbf{0.93} & 0.14 & 0.24 & \underline{0.78} & 0.45 & 0.42 \\
\textbf{Llama 2 } & \textbf{1.00} & 0.06 & 0.11 & 0.89 & \underline{0.58} & \underline{0.70} & 0.51 & \underline{0.93} & \underline{0.66} & 0.54 & 0.86 & \underline{0.66} & 0.76 & \underline{0.64} & \underline{0.61} \\

\textbf{GPT-3.5} & 0.78 &\bf 0.76 & \textbf{0.77} &\bf 0.96 &\bf 0.71 & \textbf{0.82} &\bf 0.66 & 0.90 & \textbf{0.76} & \underline{0.77} & \underline{0.92} & \textbf{0.84} &\bf 0.83 &\bf 0.80 & \textbf{0.80} \\

\hline
\end{tabular}
}
\label{tab:results-smile-data-llms}
\vspace{-2mm}
\end{table*}


\section{Experiments} 


\subsection{Benchmark Methods}
To investigate the three target tasks listed in Section III.C, we perform sentiment analysis by considering \textbf{Logistic Regression} (LR) and \textbf{Support Vector Machine} (SVM) as baseline models. 
Subsequently, we also fine-tuned \textbf{BERT} for the specific sentiment prediction task.
Additionally, we developed task-specific fine-grained prompts for Large Language Models (LLMs) to predict the sentiment labels of student responses. The four LLMs evaluated include: 
(1) \textbf{GPT-3.5} \cite{ye2023comprehensive}, 
(2) \textbf{Mistral} 7 Billion (8-bit quantization) \cite{jiang2023mistral}, 
(3) \textbf{Llama 2} 7 Billion (8-bit quantization) \cite{touvron2023llama}, and 
(4) \textbf{Orca 2} 7 Billion (8-bit quantization) \cite{mitra2023orca}. 
LLMs enable the classification of the student responses into varying levels of their satisfaction with the mental health services.

The prompt design process is iterative and data-driven to optimize the language models' performance in contextually understanding and analyzing students' survey responses.
Following the design and result analysis of the coarse prompt in Section \ref{sec:data_annotation_SMILE-College}, we develop a fine-grained prompt for sentiment prediction, which consists of the same three components as in the coarse prompt but with four fine-grained sentiment categories (Mixed) and provides specific criteria for each category (see Section \ref{sec:data_annotation_SMILE-College}).

Since LR and SVM require a training phase, we randomly split the dataset into train, development, and test with the ratios of 0.75/0.05/0.2 and report the results obtained on the test split. The same data split was used for finetuning BERT. To ensure a fair comparison, results for the four LLMs were also obtained from the same test set and provided in Table IV. Additionally, we evaluated the performance of the LLMs on the entire SMILE-College dataset (Table V and Fig. 2).

%% file: 4-Experimentation.tex

\subsection{Experimental Setup}

To evaluate the performance of the models on sentiment prediction, 
We adopt \textit{Precision}, \textit{Recall}, and \textit{F1-score} to evaluate the performance of sentiment predictions. The higher values of these metrics indicate the better performance of a model. 
We evaluated the overall performance of all sentiment categories with a weighted evaluation to handle the label imbalance issue in the data and ensure a reasonable consideration of all sentiment categories during the evaluation.
We use TensorFlow~\cite{tensorflow2015-whitepaper} and the Hugging Face library to implement various language models. Our experiments are conducted on a Nvidia Tesla V100 GPU, equipped with 51GB of RAM and 201.2GB of disk space, which has the necessary computational power. During the inference phase, we experiment with 4 or 8-bit for model quantization~\cite{jacob2017quantization}, and with temperature settings ranging from 0.1 to 0.3 to get the best results.

\subsection{Experimental Results}

\begin{figure*}[!tp]
	\centering
	\begin{subfigure}[b]{0.35\textwidth}
		\includegraphics[width=\textwidth]{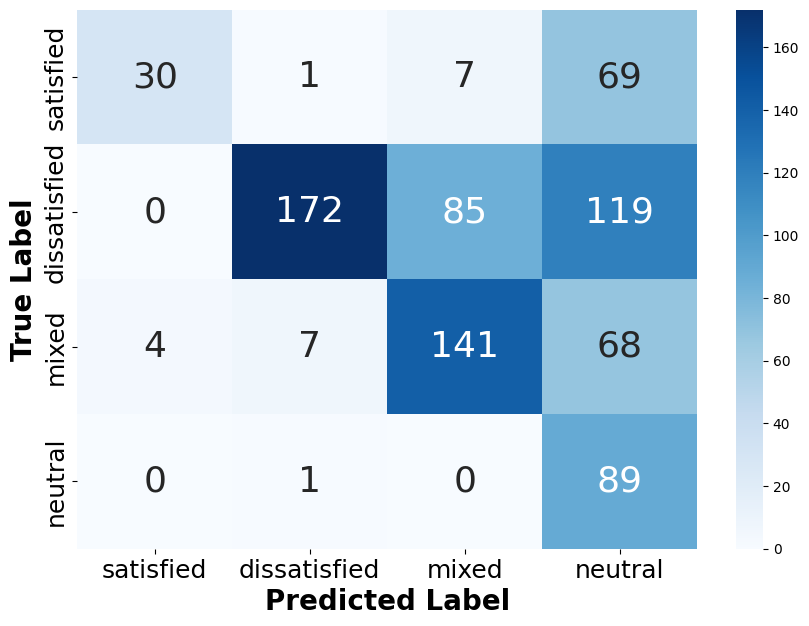}
		\caption{Mistral}
		\label{fig:attention_visualization_a}
	\end{subfigure}
        \hspace{4mm}
	\begin{subfigure}[b]{0.35\textwidth}
		\includegraphics[width=\textwidth]{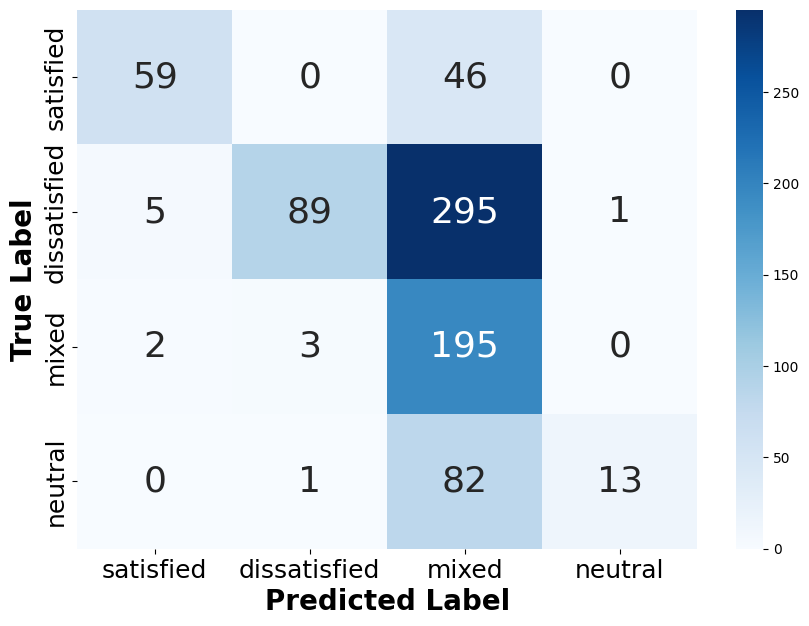}
		\caption{Orca 2}
		\label{fig:attention_visualization_b}
	\end{subfigure}
	\begin{subfigure}[b]{0.35\textwidth}
		\includegraphics[width=\textwidth]{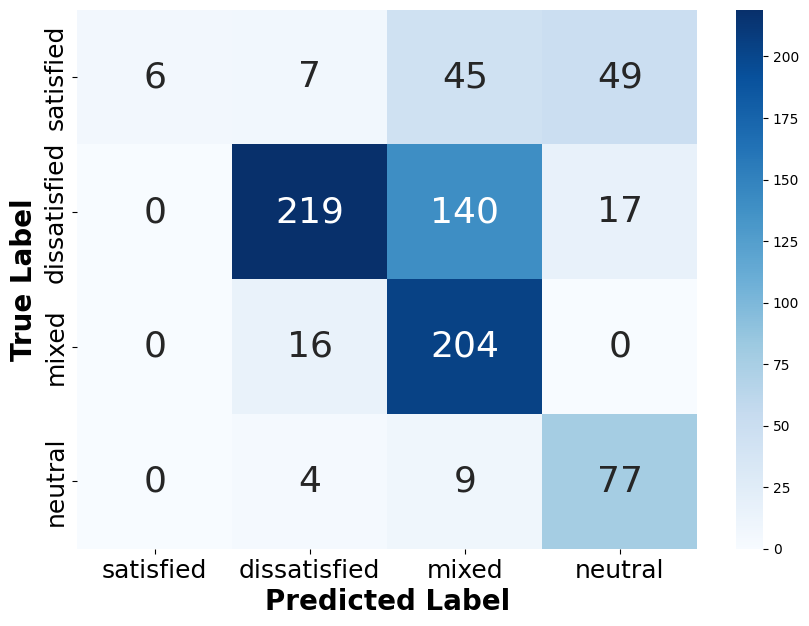}
		\caption{Llama 2}
		\label{fig:attention_visualization_c}
	\end{subfigure}
        \hspace{4mm}
	\begin{subfigure}[b]{0.35\textwidth}
		\includegraphics[width=\textwidth]{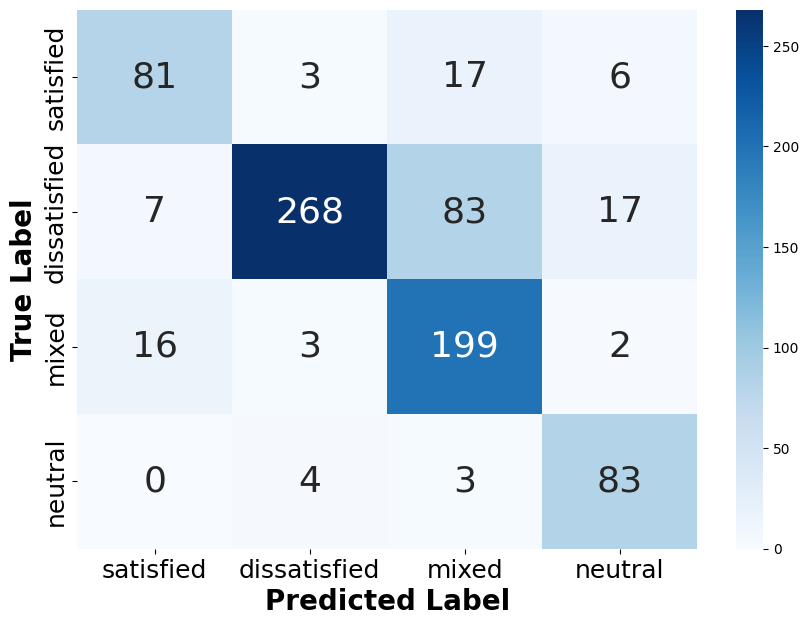}
		\caption{GPT-3.5}
		\label{fig:attention_visualization_d}
	\end{subfigure}
	\caption{Confusion metrics for the four LLMs on \textit{the entire SMILE-College dataset}.}
	\vspace{-4mm}
	\label{fig:attention_visualization}
\end{figure*}

\subsubsection{\textbf{Sentiment Prediction (\textbf{T1})}}
Table \ref{tab:results-smile-data} provides the quantitative performance comparison of different models on the test set of the SMILE-College dataset. The best performance of different models is highlighted in bold, while the second-best performance is underlined. 

We observe from Table \ref{tab:results-smile-data} that overall GPT-3.5 achieves the highest F1 score of 0.80, outperforming other models. Its large size and robust architecture allow it to deliver a balanced performance across all sentiment categories. The second-best performance is observed from the fine-tuned BERT, with an overall F1 score of 0.78. BERT consistently performs well across most categories, particularly in the Dissatisfied and Neutral categories, where it achieves an F1 score of 0.86 and 0.80, respectively. Its encoder-based architecture continues to be effective in capturing contextual relationships, leading to strong results in sentiment classification.

Interestingly, the other three LLMs, including Mistral, Orca 2, and Llama 2, while more suited for generative tasks, still deliver competitive results when compared to baselines like SVM and LR. This suggests that despite being optimized for text generation, these models exhibit a strong understanding of the contextual intricacies of mental health-related text. For instance, Mistral demonstrates strong recall in the Neutral category (1.00) and Orca 2 exhibits impressive precision in the Dissatisfied (0.95) and Satisfied (0.88) categories. However, Llama 2 underperforms significantly in the Satisfied category, where it fails to produce any meaningful results. This variability suggests that while these models grasp the overall sentiment context, their task-specific performance is not as fine-tuned or consistent as models like BERT or GPT-3.5.

The performance of the four LLM's using zero shot prompting on the entire dataset can be seen in Table \ref{tab:results-smile-data-llms}.  We observe that GPT-3.5 maintains strong performance with the highest overall F1 score, demonstrating its ability to adapt well in a zero-shot setting without the need for task-specific fine-tuning. Orca 2 demonstrates strong precision but struggles with low recall, particularly in the Dissatisfied and Neutral categories, resulting in a lower overall F1 score. Mistral excels in Neutral recall but suffers from inconsistent precision across categories. Llama 2, notably, performs better for the Satisfied category in the entire dataset (F1 = 0.11) than the test set (F1 = 0.00). Overall, the decoder-based LLMs show consistent performance across both, the test set and the entire dataset.

\subsubsection{\textbf{Prediction Error Analysis with Confusion Matrix (\textbf{T2})}}
Table \ref{tab:results-smile-data-llms} and the confusion matrices in Figure \ref{fig:attention_visualization} provide a deeper understanding of the LLM's performance variations and error patterns. Across the board, GPT-3.5 shows the most balanced distribution of errors, as reflected by its fewer misclassifications between categories. Notably, there is minimal confusion between the Satisfied and Neutral or Mixed categories, a challenge that other models face more frequently. The matrix reveals that GPT-3.5 handles the overlap between sentiments better than others.

Llama 2 demonstrates the second-best performance in LLMs, excelling in Satisfied sentiment detection with perfect precision. However, frequent misclassifications into Neutral or Mixed categories (Fig. 2(c)), highlight its struggle with recall, leading to an imbalanced performance. Despite this, Llama 2 manages a stronger performance in the Dissatisfied and Mixed categories compared to other models, showing that it can capture negative and mixed emotions fairly well, but lacks the balance needed for more diverse sentiment types.

Mistral and Orca 2 struggle with handling Mixed and Dissatisfied categories. Mistral frequently misclassifies Mixed samples as Neutral or Dissatisfied, while Orca 2 shows confusion between Mixed and Dissatisfied sentiments, though it performs well with Neutral sentiments.

\begin{table*}[!tp]
\centering
\caption{Frequency of Identified Limitations in Mental Health Services by Cluster.}
\label{table:issues}
\renewcommand\arraystretch{1.0}
\begin{tabular}{c|l|r}
\hline
\textbf{Cluster} & \textbf{Limitations}                              & \textbf{Freq} \\ \hline
1 & Quality of Counseling Services              & 157                \\ 
2 & Availability and Accessibility              & 76                \\ 
3 & Challenges in accessing the services        & 76                \\ 
4 & Awareness and Education                     & 73                \\ 
5 & Issues with Therapist Matching              & 65                \\ 
6 & Inadequacies in support, communication, community connection                   & 64                \\ 
7 & Personal Experiences and Preferences        & 52                \\ 
8 & Financial and Administrative Concerns       & 48                \\ 
9 & Diversity and Inclusivity       & 22                 \\ 
10 & Issues with Referrals and Redirection                & 13                 \\ \hline
\end{tabular}
\vspace{-4mm}
\end{table*}

\subsubsection{\textbf{Support Limitations Identification (\textbf{T3})}}
To enhance the well-being of students, it is crucial to carefully examine the areas of college mental health services that need more attention. We employ GPT-3.5 to identify and extract the limitations based on the survey responses labeled as ``Dissatisfied" in the SMILE-College dataset. 
After manual verification, we obtain the embeddings of the extracted limitations using the sentence transformer \cite{wang2020minilm} and further cluster them using K-Means \cite{5453745} to systematically categorize the limitations of college mental health services. 
The examination of each cluster's content reveals predominant themes and topics, which are systematically detailed in Table \ref{table:issues} along with the frequencies of the limitations mentioned in all the survey responses. 
Examining the dataset reveals that the most pressing issue is the quality of counseling services, with 157 mentions, followed by concerns about availability and accessibility, each cited 76 times. These findings highlight the need for colleges to improve counseling services and access. Although less frequent, issues like diversity and inclusivity and referrals also indicate areas for improvement in creating a more inclusive support system.

%% file: 5-Conclusion.tex
\vspace{-3mm}

\section{Discussions}

Working with real-world student voice survey data presents unique challenges, especially due to the unstructured and often inconsistent nature of student feedback. Data quality is entirely dependent on the respondents' willingness and seriousness to give answers. The open-ended design and subjective nature of survey questions complicated analysis with their broad range of responses. Additionally, inconsistent text generation from decoder-based LLMs made post-processing difficult, limiting the extraction of consistent insights.

Leveraging LLMs offers a significant opportunity to shed light on how mental health support structures are perceived within academic institutions. Additionally, LLMs allow for scalable analysis of subjective data and more personalized mental health interventions, helping shape data-driven policies that better meet student needs and enhance overall mental health services. The ability to highlight recurring issues can prompt institutions to make necessary revisions, improving overall mental health support systems for a more inclusive and effective approach. With this initial exploration of student sentiment on mental health support in colleges using LLMs, we hope to inspire further research into leveraging LLMs to advance mental health-related studies.

{In this work, we prioritized ethical considerations, particularly regarding student privacy and potential bias. The Student Voice Survey (SVS) Data, containing students' feedback on mental health services, was already anonymized and de-identified by College Pulse prior to annotation, ensuring privacy protection. To enhance efficiency and accuracy, we employed LLM-based annotations, which were cross-verified by human annotators from different backgrounds. This multi-layered approach minimized bias and ensured cultural relevance. Additionally, we transparently documented the role of LLMs in the annotation process and used the publicly available, vetted SVS dataset, aligning with ethical standards for privacy, fairness, and responsible AI use in mental health research.}

This work offers significant potential for advancing real-world practices in survey design and data utilization for mental health research. 
For example, the insights uncovered through the human-machine collaborative annotation process, such as the insufficient or irrelevant survey responses, and the introduction of the ``Mixed" sentiment category, underscore the critical role of well-designed survey questions in engaging participants effectively and eliciting more structured, informative responses.      
Additionally, the SMILE-College dataset's relatively small sample size poses challenges to model performance. 
Expanding the dataset with additional survey responses in future studies could enhance its robustness and generalizability. 
Future iterations of the SMILE-College dataset could also incorporate richer annotations, capturing specific issues, benefits, and emotional tone. This includes introducing more granular sentiment categories, such as differentiating dissatisfaction (e.g., service quality vs. accessibility) and satisfaction (e.g., effectiveness vs. convenience), to enable deeper analysis.


\section{Conclusions}
Mental health support in colleges and universities is crucial for fostering students' mental health awareness and well-being. However, its effectiveness is hard to evaluate due to various challenges. 
This paper utilizes student feedback from a public Student Voice Survey, employing advanced LLMs to analyze students' perceptions of college mental health support. A new SMILE-College dataset is created through human-machine collaboration for sentiment analysis. Three important tasks are investigated on the new data, including sentiment prediction, prediction error analysis, and support limitation identification. {Experiments reveal that GPT-3.5 performs the best, followed by BERT, in the sentiment prediction task. Additionally, ``Quality of Counseling Services" emerged as the most frequently identified limitations faced by students.} This data-driven approach facilitates better mental health support evaluation and decision-making.


